\documentclass{article}
\usepackage{iclr2022_workshop,times}

\PassOptionsToPackage{sort}{natbib}

\usepackage{algorithm}
\usepackage{algorithmic}
\usepackage{microtype}
\usepackage{graphicx}
\usepackage{subfigure}
\usepackage{booktabs} 
\usepackage{wrapfig}

\usepackage{mleftright}
\usepackage{nicefrac}
\usepackage{paralist}

\usepackage{hyperref}

\usepackage{amsmath}
\usepackage{amssymb}
\usepackage{mathtools}
\usepackage{amsthm}

\theoremstyle{plain}

\theoremstyle{definition}

\theoremstyle{remark}

\usepackage[textsize=tiny]{todonotes}

\newcommand{\R}{\mathbb{R}}

\DeclareMathOperator{\dist}       {d}
\DeclareMathOperator{\similarity} {s}

\DeclareMathOperator{\wasserstein}{W}

\usepackage{xspace}

\newcommand{\method}{\texttt{node2vec}\xspace}

\iclrfinalcopy
\begin{document}

\title{On the Surprising Behaviour of \method}

\author{%
  Celia Hacker\\
  Institute of Mathematics, Ecole Polytechnique Fédérale de Lausanne\\
  \texttt{celia.hacker@epfl.ch}%
  \And
  Bastian Rieck\\
  Institute of AI for Health, Helmholtz Munich and Technical University of Munich\\
  \texttt{bastian.rieck@helmholtz-muenchen.de}
}

\maketitle

\begin{abstract}
  Graph embedding techniques are a staple of modern graph learning
  research. When using embeddings for downstream tasks such as
  classification, information about their stability and robustness,
  i.e., their susceptibility to sources of noise, stochastic effects, or
  specific parameter choices, becomes increasingly important. As one of
  the most prominent graph embedding schemes, we focus on
  \method and analyse its embedding quality from multiple
  perspectives. Our findings indicate that embedding quality is unstable with
  respect to parameter choices, and we propose strategies to remedy this in practice. 
\end{abstract}

\section{Introduction}

\begin{wrapfigure}[25]{l}{0.50\linewidth}
  \centering
  \includegraphics[width=\linewidth]{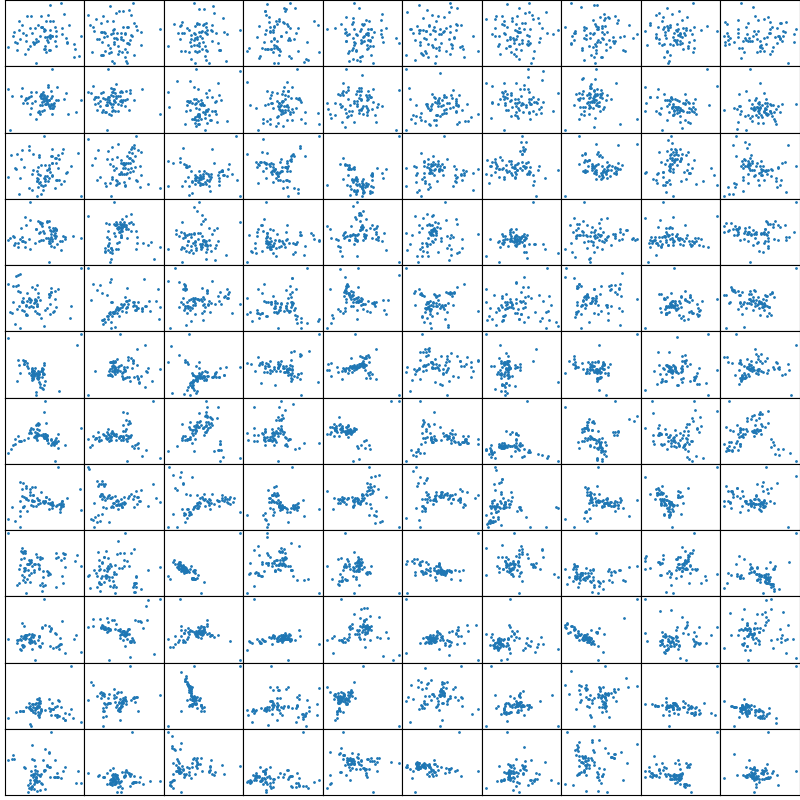}%
  \caption{%
    Different embeddings of the ``Les Mis{\'erables}'' graph for $d
    = 32$ dimensions. Each row shows embeddings that are generated from
    the \emph{same} parameter set.
    The variability between these plots is thus an indicator of
    stochastic effects in the algorithm.
    Note that this is only an illustration; in the remainder of the
    paper, we will analyse high-dimensional embeddings \emph{directly}.
  }
  \label{fig:Embeddings LM}
\end{wrapfigure}

Graph representation learning methods are used in numerous tasks, such
as graph classification, that require representing a given data set as
elements in a Euclidean space.
Many of these methods involve a certain amount of stochasticity. In
\method~\cite{node2vec}, for instance, stochasticity is present at
several stages of the algorithms, namely in the random walks themselves
and in the random initialization and optimization of the algorithm. This
algorithm also depends on a number of parameters, chosen by the
user, such as
\begin{inparaenum}[(i)]
  \item the length of random walks,
  \item the number of random walks,
  \item the dimension of the target vector space, and
  \item the neighbourhood size.
\end{inparaenum}
Hence, the resulting embeddings can vary considerably.
This is illustrated in \autoref{fig:Embeddings LM}, which depicts
low-dimensional projections of different embeddings of the
``Les Mis\'erables'' character co-occurrence network~\citep{Knuth1993TheSG}.

In practice, graph representation learning methods are often used under
the assumption that their outputs give a desired result, without
formal evidence, and any potential variability in the resulting
embeddings is not commonly studied.
We stress that this is not a unique characteristic of \method, and other
methods are also prone to it~\citep{Grohe20, ks2v}; however, in the
interest of conciseness, our paper focuses only on \method.
In \citet{hajij2021persistent}, for example, \method is applied to a graph
$G$ to obtain a map $f\colon G\to \R$. This map is then used to create
a filtration of the graph $G$ and compute persistent
homology~(see e.g.~\citet{Hensel21} for a recent survey).
In neuroscience, \method has been used in several
contexts~\citep{rosenthal, Sporns}, where the embeddings of connectomes
have been used to approximate functional connectomes from the embeddings
of the structural ones. However, in none of these applications nor in
others is it formally taken into account that the results could greatly
depend on the differences in the resulting embeddings. 
Since in principle, the outputs of the algorithm can be different, even
if the parameters are well chosen~(for instance with respect to
a specific task), it is important to understand the
stability of these methods in order to draw coherent conclusions about
the data.

Previous research \citep{Goyal_2018, ks2v} suggests that, at least within
 a range of parameters, there is a form of stability at
 a \emph{macroscopic} level. The clusters found in the point clouds
 recover almost exactly the combinatorial type of nodes of the graph or
 simplices found in the clique complex of the original graph. The
 stability exhibited at a macroscopic level motivates a more formal
 study of stability, at a more granular level
exploring local structures of the embeddings. Some existing
 approaches to the question are explained in \autoref{sec:Background}.
 In \autoref{sec:methods}, we present our own approach to this question,
 while \autoref{sec:experiments} provides the results of our
 experiments.

 \paragraph{Contributions.} We study the structural
 stability  of \method in a more formal framework by evaluating
 embeddings generated from varying sets of parameters to capture the differences
 that can be induced by different choices of parameters. We incorporate
 both ``real-world" graph data sets and well-understood random graph
 models, which will serve as null models. Overall, we find that \method
 embeddings can be unstable even for related parameter sets,
 prompting additional investigations into the stability and overall
 veracity of an embedding.

\section{Background and Related Work}\label{sec:Background}

We provide below a short reminder of the \method algorithm; more
details can be found in \citet{node2vec}.
There are several parameters to be chosen by the user, namely  the
length $L$ of random walks,  the number $N$ of random walks starting at each
node of the graph, the dimension $d$ of the embedding space, and the
neighbourhood size $C$. In the random walks one can also choose two
parameters $p,q$ which govern the random walks by
biasing them towards different behaviours. Briefly, \method~(see the
appendix\footnote{%
  See \url{https://arxiv.org/abs/2206.08252}.
}for algorithmic details), can be divided into the following main steps
\begin{compactenum}
    \item Compute a collection of truncated random walks, with biases
      $p,q$, all of length $L$, such that there are precisely $N$
      starting at each node of the graph.
    \item Using the random walks as a way to extract a measure of
      similarity between nodes, compute a representation of the nodes,
      with the paradigm that nodes that co-occur more often in random
      walks should be more likely to close by in the
      Euclidean representation. 
    \item Obtain a $d$-dimensional representation of the nodes of the
      graph reflecting the structure of the graph according to
      neighbourhoods computed through the random walks, using the
      skip-gram architecture \cite{word2vec,word2vecarxiv}
\end{compactenum} 

We want to study the stability of \method more formally. As outlined
above, there are several sources of randomness in the two algorithms,
namely in the random walks themselves, in the initialization and
training of the network.
One could take the approach of controlling stability at each step of the
algorithm in order to gauge the overall stability. The approach in this
paper will be to first understand the stability of the outputs
\emph{without} controlling the stability of each intermediate step. 
However, this question is very broad and complex. In this paper, we
propose a preliminary approach to this question by studying empirically
the effects of the change of parameters on the embeddings, using various
measures to assess the stability of the embedding at the microscopic level.

\paragraph{Related work.}
Although no formal theoretical study of stability has been provided yet,
several approaches to assess ``how good'' graph embedding methods are
have been explored, see for instance \citet{schumacher2020effects, Bonner_2017,
Xin_2019} for the study of the stability of various graph
representation learning methods. These are all empirical approaches.
Providing a theoretical framework for this question has proven to be
a difficult task given the many parameters to control, even in the
simplest cases. 
The aforementioned literature presents two
approaches to deciding whether graph embeddings are good or not are
proposed: an approach based on \emph{task
evaluation} and a more \emph{structural} or
\emph{geometric} approach.
In the task evaluation approach, the embedding method is usually
evaluated on the quality of a downstream task, such as graph
classification.
This is studied for example in \citet{schumacher2020effects}. The
drawback of such a method is that it can be specifically tested only on
labelled graphs.
At the intersection of the structural and task approach, one finds tasks
such as link and topological feature prediction~\citep{Bonner_2017,
Xin_2019}.
In the structural or geometric approach, the graph embeddings are
evaluated on ``how well" the embeddings reflect some aspects of the
graph structure. Aspects that can be evaluated include graph reconstruction, comparison of degree distributions
between that of the original and reconstructed graphs, and cluster
preservation. These are then evaluated using various measures for each aspect~\citep{Xin_2019}. Each measured aspect takes into account
different scales of the graph structure, going from a local scale where
the method is evaluated around each node, to a more global scale where
the preservation of the community structure is measured. The former was
already studied to some extent in \citet{ks2v} for
$k$-\texttt{simplex2vec} and in \citet{Goyal_2018} for \method and
similar methods. These are more related to the question of
\emph{stability} where the central question is: ``for a small variation in
the hyperparameters of the algorithm or in the graph structure, is the
variation of all criteria described above small as well?''
Some drawbacks of the evaluation methods presented in
\citet{Bonner_2017, Xin_2019} is that they evaluate the embeddings only
for a fixed choice of parameters for the embedding algorithms, thus
lacking an overview of the overall stability of the methods, omitting
the potential effects of the parameters on the embeddings.

\section{Methods}\label{sec:methods}

We consider \method as a method to produce a map~$f\colon
V\to\R^d$, i.e., a map from the set of nodes $V$ of a graph to
a $d$-dimensional vector space.
Hence, $f$ can be considered as creating a point cloud from a graph or,
equivalently, assigning to each node~$v$ in the graph a vector~$f(v) \in
\R^d$.
The advantage of this perspective is that our analysis and suggestions
apply to a wide variety of embedding schemes, including graph neural
networks.
Hence, instead of assessing the mapping~$f$ directly, we assess~$f$ in terms of
its empirical outputs.
To gauge the output of \method, we analyse its
\emph{stability} and \emph{quality}. The former term refers to the
desirable property that embeddings generated by \method should be
similar\footnote{%
  We provide a more precise notion of this similarity below.
}
to each other across different runs of the algorithm with
the same hyperparameters, whereas the latter term refers to the need for
prroducing embeddings that are useful in downstream tasks, such as the node
classification or general knowledge extraction tasks.

\subsection{Stability}\label{sec:Stability}

Our definition of stability is based on the observation that
\texttt{node2vec}, like other embedding algorithms, requires the
choice of several parameters. Moreover, due to the way the embeddings
are constructed, there is a degree of stochasticity involved---the
initialisation of the optimiser, for instance, has an impact on the
output. We consider an embedding scheme to be \emph{stable} if its
outputs are ``reasonably close'' for parameter sets that are
``reasonably close.'' More formally, assuming that $X$ and $Y$ refer to
embeddings of the same graph, obtained from parameter sets $\Theta_X$
and $\Theta_Y$, respectively, we want
\begin{equation}
    d(X, Y) \propto d(\Theta_X, \Theta_Y).
    \label{eq:Stability}
\end{equation}
In other words, the distance between embeddings $X$ and $Y$ should be
proportional to the distance between their parameter sets, for some
appropriate choice of distance in each case.\footnote{%
  Here, we use a simple distance based on the Hamming metric between
  parameter sets to obtain a consistent ordering between parameter sets.
  We leave a more detailed analysis of this equation for future work.
}
We argue that
this is a natural desirable property of an algorithm because it implies
that all conclusions drawn from a specific embedding change in
a predictable way as one chooses different parameters.

To assess the stability of embeddings, we will make use of two
measures for comparing point clouds. First, the \emph{Hausdorff
distance} between two point clouds is defined as
\begin{equation}
    d_{\mathrm {H} }(X,Y)=\max \left\{\,\sup _{x\in X}d(x,Y),\,\sup _{y\in Y}d(X,y)\,\right\},
\end{equation}
with $d(x, Y) := \inf _{y\in Y}d(x,y)$, i.e., the minimum distance from
$x$ to the point cloud~$Y$, and vice versa for $d(X, y)$. The Hausdorff
distance is easy to calculate but relies on the relative positions of the point clouds $X$ and $Y$ in $\R^d$. Hence, the distance is sensitive to the position of the two point
clouds. 

To account for small changes in their relative positions,\footnote{%
  A more advanced option would be to \emph{register} point clouds first
  in order to be invariant to rotations and translations. We leave this
  for future work.
}
we will also work with the \emph{Wasserstein distance} of order~2,
\begin{equation}
    \wasserstein_2(X, Y) := \mleft(\inf_{\eta\colon X \to
    Y}\sum_{x\in{}X}\|x-\eta(x)\|_2^2\mright)^{\frac{1}{2}},
\end{equation}
ranging over all bijections $\eta\colon X \to Y$ between the two point
clouds, where we compare embeddings computed from the same graph.

Given point clouds $X_1, X_2, \dots, X_n$, arising from the same
parameter set~$\Theta_X$, we can calculate their \emph{inter-group distances} via
the two metrics defined above. If \method is \emph{stable} for
$\Theta_X$, we expect low variance of such distances. Moreover,
following \autoref{eq:Stability}, we permit that the
distributions of inter-group distances become markedly different as the
respective parameter sets become increasingly different. For instance,
increasing the number of random walks from $L$ to $L + 1$ may result in
distributions that are close~(in the statistical sense), whereas going
from $L$ to $L' \gg L$ walks could result in distributions that are
dissimilar to each other.

\subsection{Quality}\label{sec:Quality}

The assessment of the quality of an embedding is distinct from its
stability. For instance, \method could be replaced by a simple map
that assigns each graph to the zero vector. Such an embedding would
be perfectly stable but its output is virtually useless in
practice.
In the context of graph learning, embedding quality is often measured in
terms of predictive performance for a specific task, e.g., how well the
embedding serves to preserve important nodes or communities. Since we
want to cover situations in which we evaluate \method without a specific
ground truth, we need \emph{task-independent quality measures}.

\subsubsection{Link distribution comparison}
Inspired by \citet{Xin_2019}, we first measure the dissimilarity between
the link distributions of a graph and a link distribution computed using
its embedding.
Letting $A$ refer to the adjacency matrix of a graph~$G = (V, E)$, we first obtain
the \emph{observed link distribution} $P^G$ as
$P^G_{ij} := \nicefrac{A_{ij}}{|E|}$.
$P^G$ provides the probability of an edge $(i,j) \in E$, under the
assumption that edges are uniformly distributed.
In an embedding~$X$ of~$G$, we do not have access to observed edges.
However, we may use a similarity measure~$\similarity(\cdot, \cdot)$
between the representations of individual nodes~$v_i, v_j$, giving rise
to an \emph{empirical link distribution}~\cite{Xin_2019} via
\begin{equation}
  P^X_{ij} := \frac{1}{1 + \exp\mleft(-\similarity\mleft(x_i, x_j\mright)\mright)},
\end{equation} 
with $x_i, x_j$ referring to the representations of vertices~$v_i, v_j$,
respectively. Notice that this distribution requires normalisation to
constitute a proper probability distribution.
There are numerous choices for realising $\similarity(x_i, x_j)$ in
practice; following \citet{Xin_2019}, we pick the standard Euclidean
inner product of $x_i, x_j$, i.e., $\similarity\mleft(x_i, x_j\mright)
:= x_i^{\top} x_j$, also known as a linear kernel.
Since \method incorporates a similar objective---its training involves
minimising inner product expressions---this formulation has proven
useful to assess the output of embedding algorithms~\cite{Grohe20}.
This formulation provides us with a suitable quality assessment
strategy: taking any measure of statistical distance~$\dist_S(\cdot,
\cdot)$, we use $\dist_S\mleft(P^G, P^X\mright)$ as an indicator of the
quality of an embedding~$X$. This has the advantage of being immediately
interpretable: lower values indicate a better ``fit,'' in the sense that
the resulting embedding preserves the link distribution of a graph, thus
permitting a reconstruction.

To implement $\dist_S(\cdot, \cdot)$, we use the \emph{Wasserstein
distance}; while \citet{Xin_2019} suggest using a KL divergence, we
prefer the Wasserstein distance because it constitutes a proper metric
between probability distributions, satisfying in particular symmetry and
the triangle inequality, two properties that are indispensable when
comparing different embeddings across parameter sets.
One advantageous property of this quality assessment procedure is that
it is \emph{independent} of the dimensionality of the embeddings; we
can thus compare the distributions across all parameter sets.

\subsubsection{Link reconstruction}
%
As a second measure for assessing the quality of embeddings, we consider the
previously-defined similarity scores between nodes, i.e.,
$\similarity\mleft(x_i, x_j\mright) := x_i^{\top} x_j$,
to be a score of classifier that aims to predict the existence of an
edge. Writing $\similarity_{ij}$ to denote $\similarity\mleft(x_i,
x_j\mright)$, we consider superlevel sets of the form
$E_{\lambda} := \{ (i,j ) \mid \similarity_{ij} \geq
\lambda\}$, for $\lambda \in \R$. For each choice of $\lambda$, we treat
$E_{\lambda}$ as the set of all edges predicted by the classifier.
Ranging over all possible choices for~$\lambda$, we calculate
a contingency table containing the number of true positives~(TP; edges that
are correctly predicted), true negatives~(TN; non-edges that are correctly
predicted, i.e., edges that are \emph{not} part of a specific
$E_{\lambda}$), false positives~(FP), and false negatives~(FN), respectively. This
permits us to calculate \emph{precision--recall} values, and we take the
area under the precision--recall curve~(AUPRC) to be a measure of the
quality of an embedding~\citep{Fawcett06}.

\paragraph{AUROC.} Alternatively, we could use the TP and FP values to calculate
a \emph{receiver--operating} characteristic curve and use its area,
known as AUC or AUROC, as a measure~\citep{Fawcett06}.
This is more complicated to interpret in most cases, though, since AUROC
assumes a binary classification task in which two classes have roughly
similar sizes. However, most real-life graphs are
highly-imbalanced: out of all the \emph{potential} edges, only
a fraction of them ever exists; a classifier that never predicts any
edges based on an embedding would thus score high in terms of AUROC.

\section{Experiments}\label{sec:experiments}

We run \method on several different models of graphs on a grid of
parameters, using the co-occurrence network of ``Les
Misérables''~\citep{Knuth1993TheSG},
as well as stochastic block models~(SBMs), exhibiting well-defined
community structures.
To prevent additional stochastic effects, we focus on embedding
a \emph{single} instance of the SBM graphs only~(using a fixed random
seed for the generation of the graph), as opposed to working with data
sets of multiple graphs.\footnote{
  We performed similar analyses for Erd\"{o}s-R\'enyi graphs,
  but found no substantial differences to the results presented here.
}

\paragraph{Implementation.}
%
We implemented our analysis in Python, using \texttt{scikit-learn} and
\texttt{pytorch}; our code is publicly available under a~3-Clause BSD
License.\footnote{%
  \url{https://github.com/aidos-lab/node2vec-surprises}
}

\paragraph{Experimental setup.}
%
The first stochastic block model~(\texttt{sbm2}) has two
blocks of $100$ nodes and intra-block probability~$0.2$ and
inter-block probability~$0.8$. The second stochastic block
model~(\texttt{sbm3}) has three
blocks of $100$ nodes with intra-block probabilities $0.8$ and inter-block
probability of $0.2$.
For each of these graphs, we run \method \ 10 times for each combination
of parameters, to assess the variance of the algorithm. Our training
procedure uses ``early stopping'' based on the loss, and we used the
following hyperparameter grid:
$L\in \{5, 10, 20\}$, $d\in \{16, 32,
64\}$, and $C\in \{5, 10\}$, $q \in \{1, 2\}$. Similar to
\citet{node2vec}, we use a fixed number of $N = 10$ random walks
to limit the complexity of the problem.\footnote{%
  We consider our work to be a preliminary study of the stability of
  \method and related methods; future work could tackle even more
  complex hyperparameter grids.
}
In total, this parameter grid provides us with an in-depth view of the 
performance of \method. 

\paragraph{Pre-processing.} To simplify the comparisons of different
embeddings from a same feature space $\R^d$, we normalise each point cloud so that its diameter is~$1$. 
This has no consequences for quality assessment, but ensures that point
clouds remain comparable for the stability analysis. 

\subsection{Analysis of ``Les Mis\'erables''}

Following the original publication~\cite{node2vec}, we first analysed
the ``Les Mis\'erables'' character co-occurrence network.
\autoref{fig:Embeddings LM} depicts different embeddings, obtained by
applying Principal Component Analysis~(PCA) to the node embeddings generated
by \method. Each point in such a plot corresponds to
a \emph{single} node of the graph.
Already in these low-dimensional plots, a large degree of variability
becomes apparent---the resulting point clouds have different
appearances, even if we only consider runs with the \emph{same}
parameter set, i.e., all point clouds of a single row of
\autoref{fig:Embeddings LM}; we acknowledge that variations in the
appearance of embeddings can also be a consequence of PCA. However, the
illustration mimics the general usage of \method in a data analysis
context, where embedding algorithms are typically used to acquire
additional insights or solve data analysis tasks.

Subsequently, we will proceed with direct analyses of the
\emph{high-dimensional embeddings}, excluding additional dimensionality
reduction algorithms to ensure that we can assess the stability and quality 
of \method without adding a source of potential noise.
Knowing that embeddings of high-dimensional data sets are prone to
incorrect interpretations, we now turn towards a more principled
analysis of the individual point clouds, according to the measures
outlined in \autoref{sec:Stability} and \autoref{sec:Quality}.

\begin{figure}[tb]
  \centering
  \subfigure[Hausdorff distance\label{sfig:Hausdorff distance LM}]{%
    \includegraphics[width=0.50\linewidth]{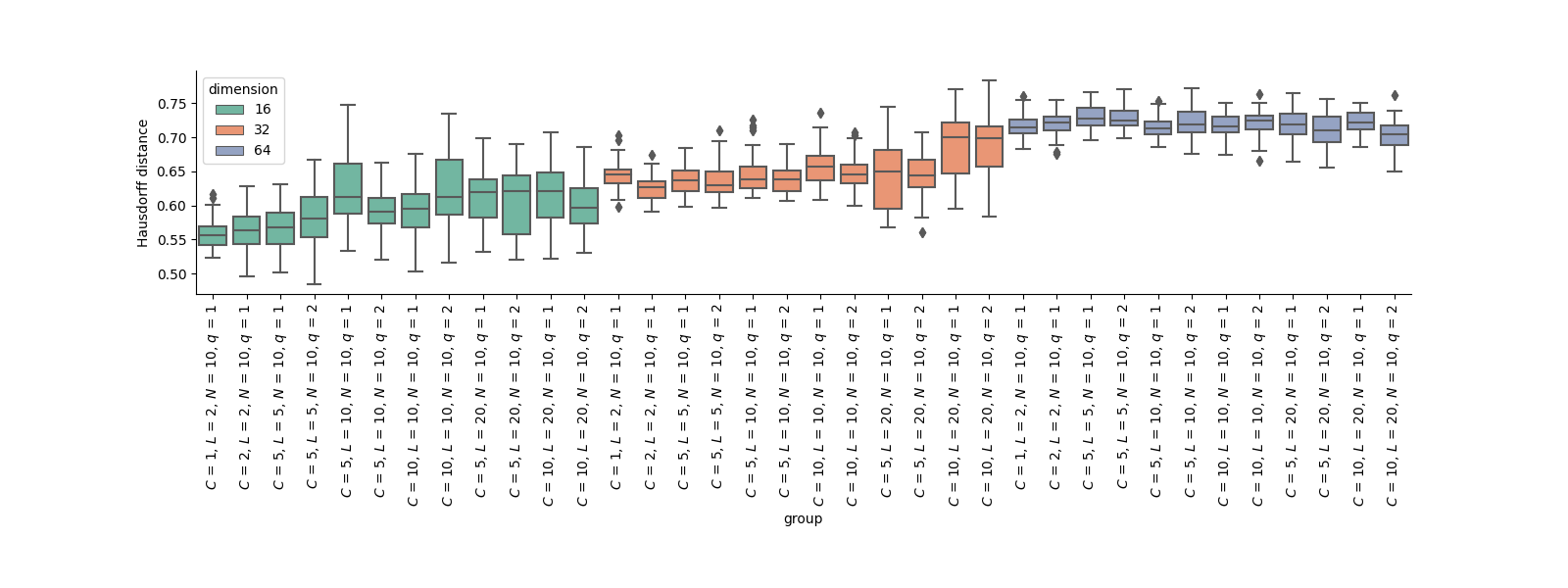}%
  }%
  \subfigure[Wasserstein distance\label{sfig:Wasserstein distance LM}]{%
    \includegraphics[width=0.50\linewidth]{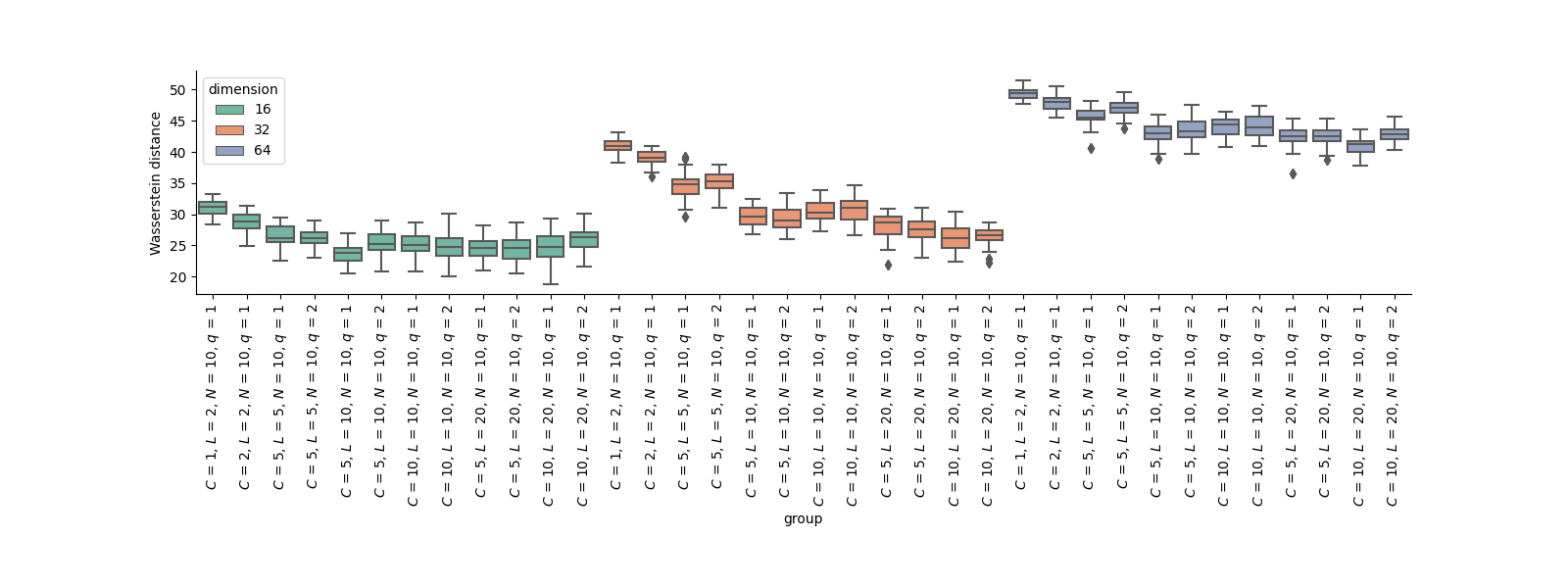}%
  }%
  \caption{%
    Distances between embeddings arising from
    a fixed parameter set for the ``Les Mis{\'erables}'' graph. The
    distributions are sorted according their parameter set and
    colour-coded according to the dimension of the respective embedding.
    Each boxplot summarises the distribution of pairwise distances with
    respect to a specific set of parameters~(shown on the
    \mbox{$x$-axis} as labels).
  }
  \label{fig:Distance distributions LM}
\end{figure}

\paragraph{Stability.}
%
\autoref{fig:Distance distributions LM} depicts distance distributions
for this data set. The Hausdorff distance exhibits a larger variance,
even \emph{within} a set of hyperparameters, i.e., over different
repetitions of the same embedding. The Wasserstein distance, by
contrast, exhibits smaller variance, hinting at its improved ability to
account for some---not all---transformations of the embeddings that are
due to stochasticity.
Using a Wilcoxon signed-rank test~\citep{Wilcoxon45}, we find that $\approx
77\%$~[$\approx90\%$] of all pairwise comparisons between hyperparameter
groups are found to be statistically significantly different from each
other for the Hausdorff~[Wasserstein] distance~($p = 0.05$ with Bonferroni
correction). This finding indicates
that even if hyperparameter sets differ only by a single value, 
\begin{inparaenum}[(i)]
  \item the resulting distribution of distances between individual
    embeddings can be substantially different, and
  \item these differences are pronounced enough to be detected by
    standard tests of statistical significance.
\end{inparaenum}
%

\begin{figure*}
  \centering
  \subfigure[Link distribution~($\wasserstein_2$)]{%
    \includegraphics[width=0.50\linewidth]{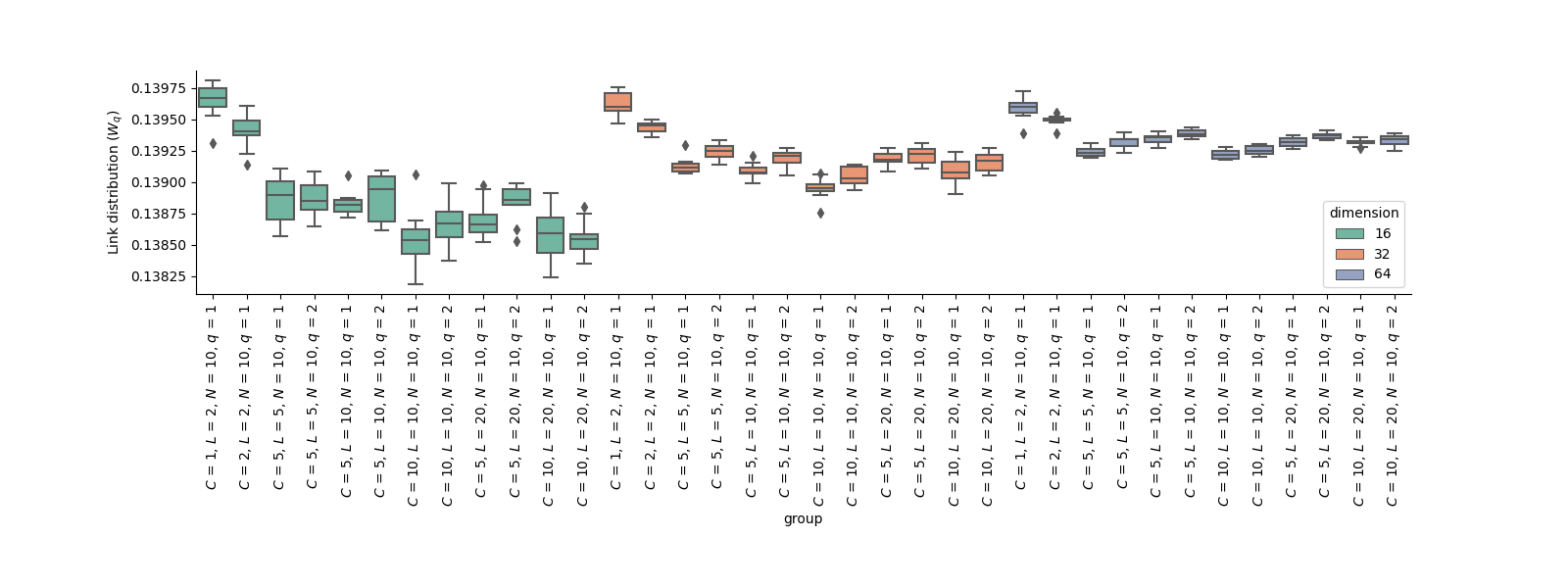}%
  }%
  \subfigure[Link reconstruction~(AUPRC)]{%
    \includegraphics[width=0.50\linewidth]{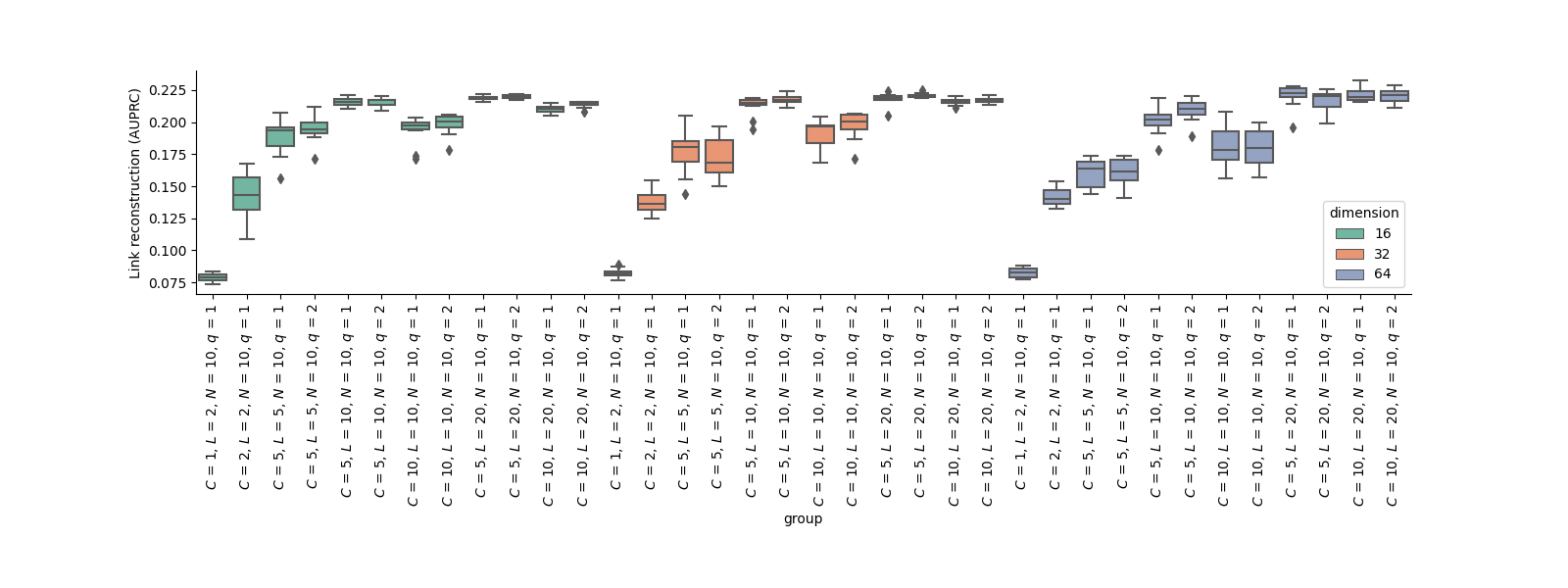}%
  }
  \caption{%
    Quality assessment of ``Les Mis{\'erables}'' graph embeddings. 
    For the link distribution comparison plot~(top), we show the 
    distribution of Wasserstein distances between the original link
    distribution of the graph and the empirical link distribution of its
    embeddings. Lower values indicate a better fit.
    For the link reconstruction~(bottom), we show the AUPRC obtained
    from reconstructing edges in the graph based on pairwise scores
    between embedded vectors. Higher values are desirable.
  }
  \label{fig:Quality LM}
\end{figure*}

\paragraph{Quality.}
%
The experiments mentioned above illustrate various ways in which the
embeddings appear to be unstable. To form a clearer picture of this
problem, we also assessed quality by aiming to reconstruct the edges in
the graph.
\autoref{fig:Quality LM} depicts the results.
We observe that there is a large degree of variance even for a single
parameter set. Somewhat surprisingly, some parameter sets achieve very
low AUPRC values, failing to go well beyond random
predictions~(AUPRC~$\approx 0.09$). The variance for certain choices of
parameters---regardless of embedding dimensionality---is still high,
indicating that even a careful selection of data-appropriate parameters
does not necessarily lead to consistently faithful embeddings.

\subsection{Stochastic block models}

\begin{figure*}
  \centering
  \subfigure[\texttt{sbm2}]{%
    \includegraphics[width=0.50\linewidth]{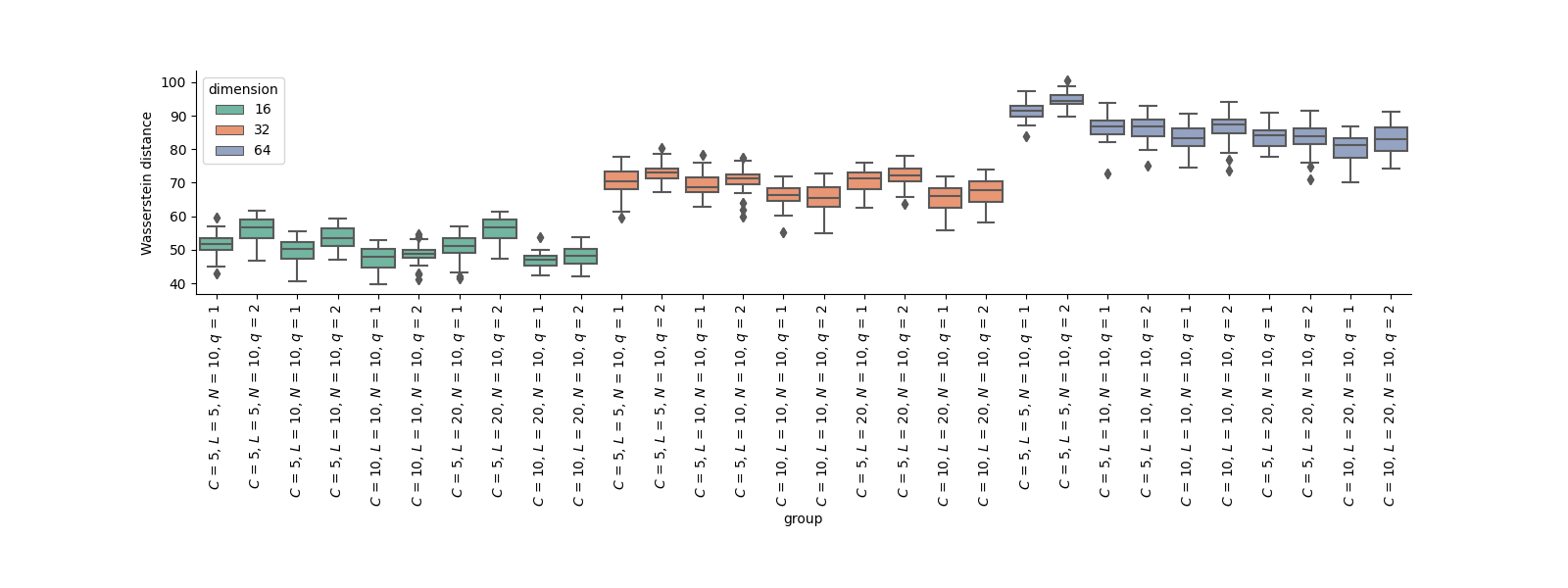}%
  }%
  \subfigure[\texttt{sbm3}]{%
    \includegraphics[width=0.50\linewidth]{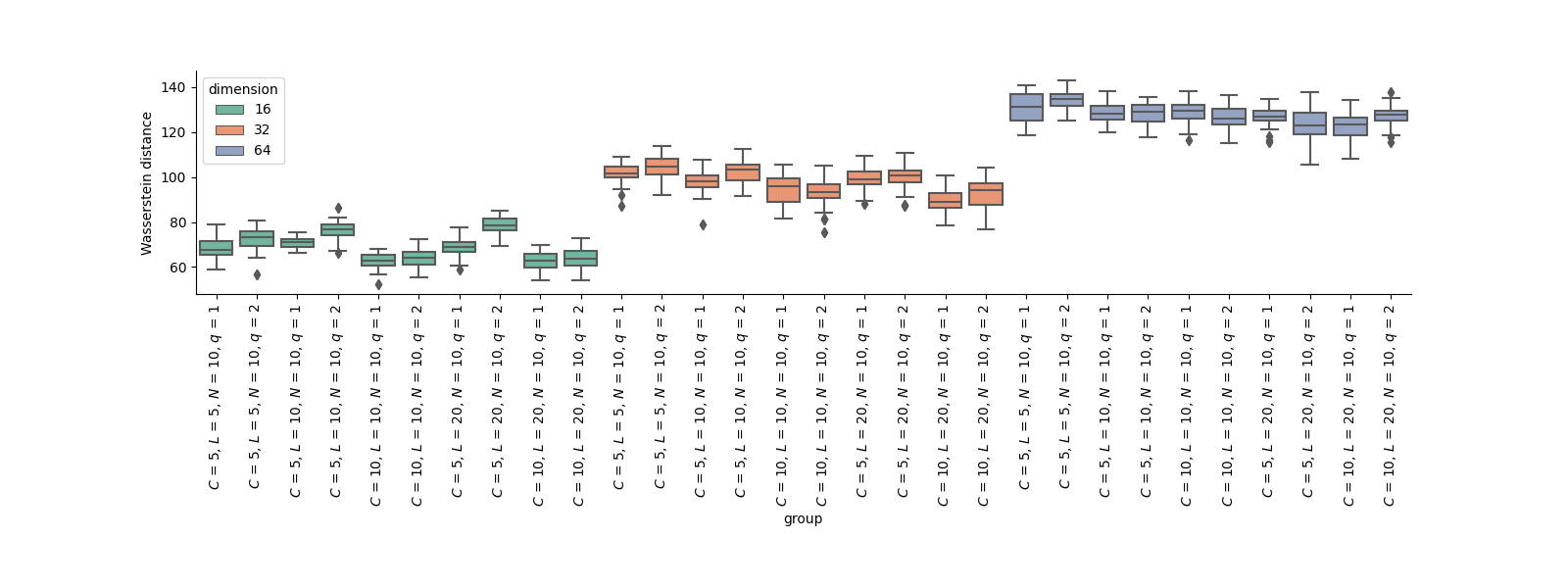}%
  }%
  \caption{%
    Wasserstein distance distributions of \texttt{sbm2} and
    \texttt{sbm3} graphs. Each boxplot summarises the distribution of
    pairwise distances with respect to a specific set of
    parameters~(shown on the \mbox{$x$-axis} as labels).
  }
  \label{fig:Stability sbm}
\end{figure*}

We subject two different graphs arising from a stochastic block model
formulation to the analysis described above; \texttt{sbm2} has
two communities, whereas \texttt{sbm3} has three.
\autoref{fig:Stability sbm} shows Wasserstein distance distributions to
assess the quality of the embeddings. Similar to the ``Les Mis{\'erables}''
graph, $\approx 91\%$ of all distributions are statistically
significantly different from each other, despite being obtained via
highly similar parameter sets.
The quality of the embeddings is highly-varying. As \autoref{fig:Quality
sbm} shows, even small changes in the hyperparameters can result in
large changes with respect to AUPRC. Moreover, it is notable that the
\emph{improvements} in terms of AUPRC over a random classifier are lower
than for the ``Les Mis{\'erables}'' graph, despite SBMs having
a relatively simple graph structure.

\begin{figure*}
  \centering
  \subfigure[\texttt{sbm2}]{%
    \includegraphics[width=0.50\linewidth]{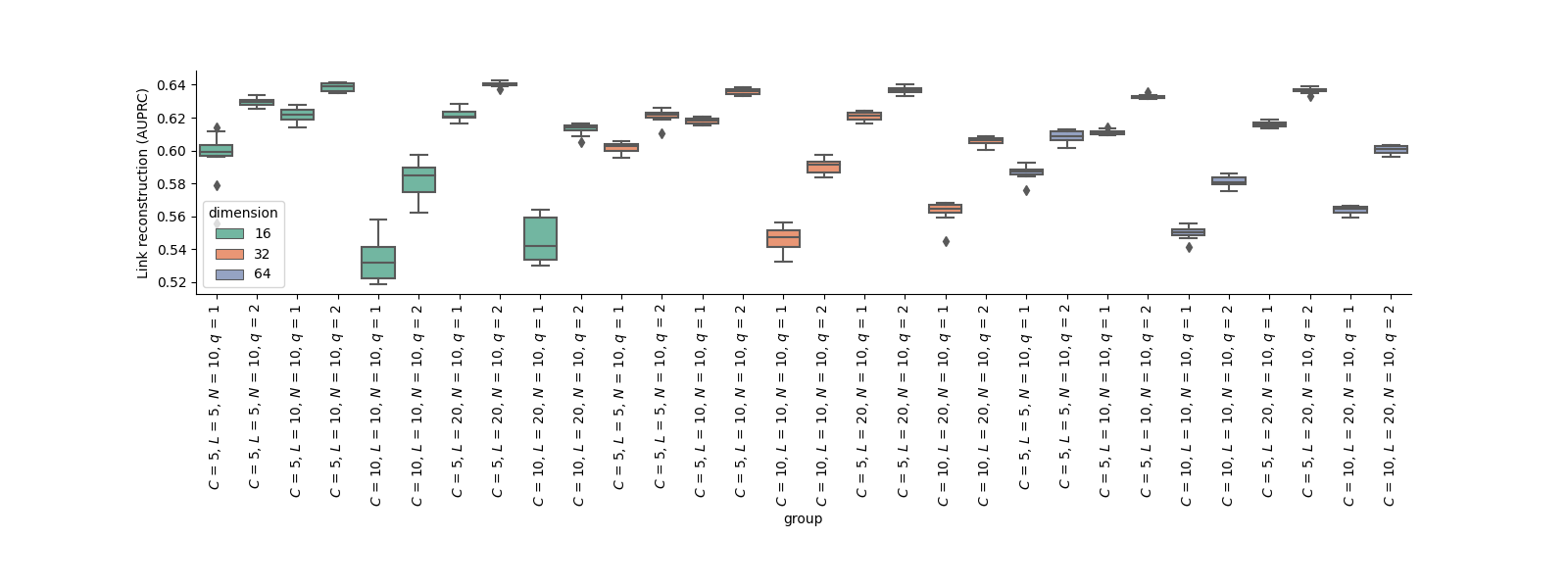}%
  }%
  \subfigure[\texttt{sbm3}]{%
    \includegraphics[width=0.50\linewidth]{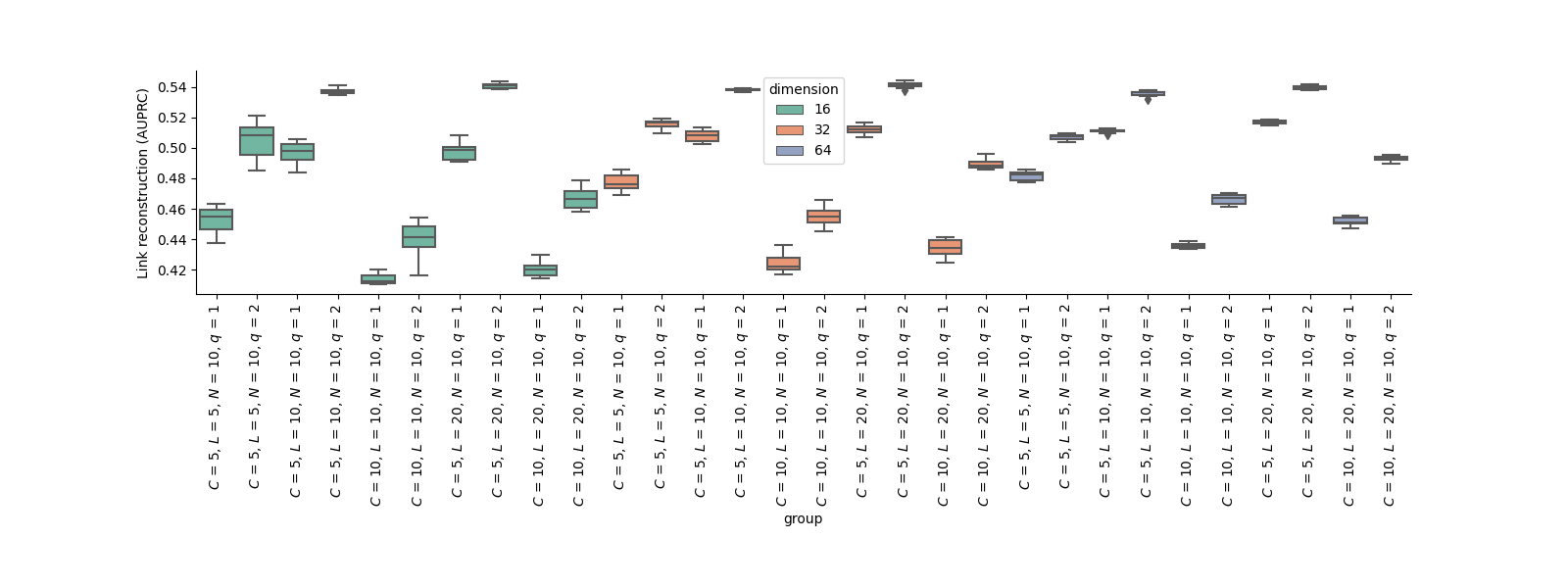}%
  }%
  \caption{%
    Link reconstruction AUPRC values for the \texttt{sbm2} and
    \texttt{sbm3} graphs. The values of a random classifier are at
    $0.50$ and $0.40$, respectively. Higher values are desirable.
  }
  \label{fig:Quality sbm}
\end{figure*}

\section{Discussion \& Conclusion}

Our experiments indicate that \method embeddings can be unstable
regardless of parameter combinations. Even if embeddings arise from the
same set of hyperparameters, there can be a large degree of variance in
terms of both stability and quality~(which we measured in terms of
reconstructing edges correctly). Our quality measure is arguably
strict in the sense that a reconstruction of \emph{neighbourhoods} would
be sufficient for many graph learning tasks. Nevertheless, the high
degree of variance~(in terms of AUPRC) and the large changes as
parameters are being adjusted, raises some concerns about the
reliability or faithfulness of such embeddings. Since \method is geared
towards producing vector representations of structured objects, there
is a need for a better understanding of how hyperparameters are to be
picked in practice. With a specific task---such as node
classification---in mind, standard hyperparameter training strategies
can be used. Our analyses indicate that care should be taken when
\method, or other vectorisation strategies, are used in an
\emph{exploratory data analysis}~(EDA) setting, i.e., for generating
hypotheses or understanding patterns underlying a data set. In such
scenarios, we caution practitioners to rely on individual embeddings.

\paragraph{Strategies.}
%
To use embeddings for EDA, we suggest repeating runs with the same
parameter set and studying the respective quality and stability
measures. While link reconstruction, for instance, might not be the
ultimate goal, low-variance distributions with high mean AUPRC values
are indicative of hyperparameters that are capable of gleaning at least
\emph{some} structural information of a graph. In addition,
practitioners might experiment with pooling strategies, i.e., averaging
individual embeddings~(potentially after aligning them with a Procrustes
analysis~\citep{Kendall89} or a registration of distributions~\citep{Tang17},
for instance). Using a single embedding to draw conclusions
about a graph should be avoided.

\paragraph{Future work.} 
Our work paves the way towards a better understanding of embedding
algorithms such as \method. For the future, we envision the development
of a more theoretical framework in which to analyse such algorithms and their outputs.
We hope that our paper opens up the discussion of this broad topic,
which might not have a simple answer.  
A better understanding of the behaviour of such methods will hopefully
also result in a principled way for choosing parameters so that they are
guaranteed to result in stable, faithful, and trustworthy embeddings,
which reflect properties of the graphs under
consideration~(such as the diameter, clustering coefficient, and other
properties). 

In this context, developing novel stability and quality measures based
on e.g., persistent homology~(PH), a method to obtain multi-scale topological
features of point clouds~\citep{Hensel21}, constitutes an interesting research
direction, PH having already demonstrated its expressivity in other data science
contexts such as dimensionality reduction~\citep{Rieck15b} and
clustering~\citep{Rieck16a}.

On a theoretical level, our analysis raises questions about the
behaviour of truncated random walks. While the behaviour of random walks
is well-understood~\citep{Chung1997} in the infinite time limit, not much
is known about the behaviour of truncated random walks. A better
understanding of the short-term behaviour or random walks would be
beneficial to the study of \method stability. 
Moreover, these questions also apply to $k$-\texttt{simplex2vec} and in
a broader context also to other representation learning methods for
graphs, as well as simplicial and cell complexes, which have recently
started receiving more attention in the graph learning
community~\citep{Bodnar21a, Bodnar21b}.

\section*{Acknowledgements}
C.H.\ is supported by NCCR-Synapsy Phase-3 SNSF grant number 51NF40-185897.

\clearpage

\bibliography{Biblio}
\bibliographystyle{iclr2022_workshop}

\newpage
\appendix
\onecolumn

\section{A Short Summary of \texttt{node2vec}}

The algorithm takes several hyperparameters, the walk length, number of
walks, and dimension of the embedding space. Using a collection of
random walks exploring the entire graph, \method produces
a representation of a graph in $\R^d$ by optimizing a similarity
measure, the standard dot product $\similarity(x_i, x_j)$, between all
pairs of embedding vectors $x_i,x_j\in\R^d$ to reflect the frequency at
which the corresponding nodes $v_i,v_j\in V$ co-occur in the collection
of random walks.

In Algorithm \ref{alg:node2vec} we recall the main steps of \method.
\begin{algorithm}
\caption{node2vec}\label{alg:node2vec}
\begin{algorithmic} 
\STATE {\bfseries Input: } Graph $G$, dimension of feature space $d$, walks per node $N$, walk length $L$, neighbourhood size $C$, $p,q$ parameters for biased random walks
\STATE {\bfseries Output: }$F: V \rightarrow \R^d$ $d$-dimensional representation of $G$
\vspace{10pt}
\STATE Walks = [ ]
\FOR{$v \in V$}
\FOR{$n=1$ {\bfseries to} $N$}
\STATE walk = [$v$]
\FOR{$l=1$ {\bfseries to} $L$} 
\STATE $w$ = RandomWalkStep($G,v$, $p$,$q$)
\STATE Append(walk, $w$)
\ENDFOR
\STATE Append(Walks, walk)
\ENDFOR
\ENDFOR
\STATE	 $F$ = SkipGram(Walks, $d$, $C$ )
\STATE {\bfseries Return} $F$
\end{algorithmic}
\end{algorithm}

\end{document}